\documentclass[letterpaper, 10 pt, conference]{ieeeconf}  

\IEEEoverridecommandlockouts                              

\usepackage[T1]{fontenc}

\usepackage{amsfonts}	
\usepackage{amsmath}	
\usepackage{amssymb}    
\usepackage{siunitx}
\usepackage{xfrac}    
\usepackage{pifont}   

\usepackage{booktabs}
\usepackage{makecell}  
\usepackage[flushleft]{threeparttable}  
\usepackage{multirow}
\usepackage{rotating}

\usepackage{xspace}    
\usepackage[usenames,dvipsnames]{xcolor}    
\usepackage{colortbl}
\let\labelindent\relax
\usepackage[inline]{enumitem} 
\usepackage{graphicx} 
\usepackage{microtype}
\usepackage{cite}
\usepackage{flushend}
\usepackage{xcolor}
\usepackage{svg}

\makeatletter
\let\NAT@parse\undefined
\makeatother

\usepackage{url}

\usepackage[pdfencoding=auto, colorlinks=true]{hyperref} 
\usepackage[hang,flushmargin]{footmisc}

\usepackage[capitalize]{cleveref}
\crefname{section}{Sec.}{Secs.}
\Crefname{section}{Section}{Sections}
\Crefname{table}{Table}{Tables}
\crefname{table}{Tab.}{Tabs.}

\begin{document}

\newcommand{\myfunding}{This work has been supported by projects PID2021-126623OB-I00 and PID2024-161576OB-I00, funded by MCIN/AEI/10.13039/501100011033 and co-funded by the European Regional Development Fund (ERDF, “A way of making Europe”), by project PLEC2023-010343 (INARTRANS 4.0) funded by MCIN/AEI/10.13039/501100011033, and by the R\&D program TEC-2024/TEC-62 (iRoboCity2030-CM) and ELLIS Unit Madrid, granted by the Community of Madrid.}

\title{\LARGE \bf
CaR1: A Multi-Modal Baseline for BEV Vehicle Segmentation via Camera-Radar Fusion
}



\author{
Santiago Montiel-Marín,
Angel Llamazares,
Miguel Antunes-García, \\
Fabio Sánchez-García,
and Luis M. Bergasa
\thanks{All authors are with the Department of Electronics. University of Alcalá, Community of Madrid, Spain.}%
\thanks{\myfunding}%
}

\maketitle
\thispagestyle{empty}
\pagestyle{empty}


\begin{abstract}
    Camera-radar fusion offers a robust and cost-effective alternative to LiDAR-based autonomous driving systems by combining complementary sensing capabilities: cameras provide rich semantic cues but unreliable depth, while radar delivers sparse yet reliable position and motion information. We introduce CaR1, a novel camera-radar fusion architecture for BEV vehicle segmentation. Built upon BEVFusion, our approach incorporates a grid-wise radar encoding that discretizes point clouds into structured BEV features and an adaptive fusion mechanism that dynamically balances sensor contributions. Experiments on nuScenes demonstrate competitive segmentation performance (57.6 IoU), on par with state-of-the-art methods. Code is publicly available \href{https://www.github.com/santimontiel/car1}{online}.
\end{abstract}


\section{Introduction}

Integrating accurate perception systems of 3D surroundings is fundamental for autonomous driving (AD), where cameras, LiDAR, and radar are the primary sensing modalities. In the field of multimodal perception, camera-LiDAR fusion has been extensively researched; however, the high cost and weather sensitivity limit its large-scale deployment. The recent development of vision-centric architectures has led to efforts in camera-radar fusion, as these sensors offer complementary characteristics at a lower cost. Cameras provide pixel-wise spatial and semantic information, but depth estimation remains ill-posed. Radar, in contrast, provides reliable distance and velocity measurements and remains robust under adverse weather, yet suffers from sparsity and low angular resolution.

A central challenge in fusing both modalities lies in the heteregeneous nature of sensor data, known as \textit{view disparity}. As cameras deliver dense images and radar measurements are unstructured and sparse point clouds, fusion is not possible without a common framework. Fusing in a unified Bird’s-Eye View (BEV) representation has become standard practice, achieving more accurate and robust perception for downstream tasks such as detection, segmentation, and motion prediction. Several strategies have emerged to project radar representations to BEV: point-based processing, handcrafted occupancy grids, or low-level range-azimuth maps, each with trade-offs in spatial resolution and computational cost. In addition, once modalities are represented in a shared space, effective fusion becomes critical. Rigid fusion strategies often fail under adverse weather (rain, snow, fog) or sensor-specific limitations such as occlusions and blind spots. To overcome this, our adaptive fusion mechanism dynamically balances the contribution of each modality, ensuring that perception performance remains robust even when one sensor degrades.

The main contributions of this report are twofold: to tackle radar projection to BEV, a grid-wise feature encoding method for radar point clouds; to weigh the impact of both modalities, an adaptive fusion strategy. We introduce \textbf{CaR1}, a novel architecture for BEV vehicle segmentation. It leverages a vision encoder with a geometric-based transform to project camera features into the BEV space and a point-based encoder that discretizes sparse radar signals into a grid representation. The resulting BEV features are unified through an adaptive fuser that weights the representation of each modality. Finally, an Attention U-Net decoder and a segmentation head generate the BEV vehicle maps. Experimentation on the nuScenes \cite{caesar2020nuscenes} dataset validates our approach, which effectively integrates complementary information from both modalities.

\begin{figure}[t]
  \centering
  \includegraphics[width=0.75\linewidth]{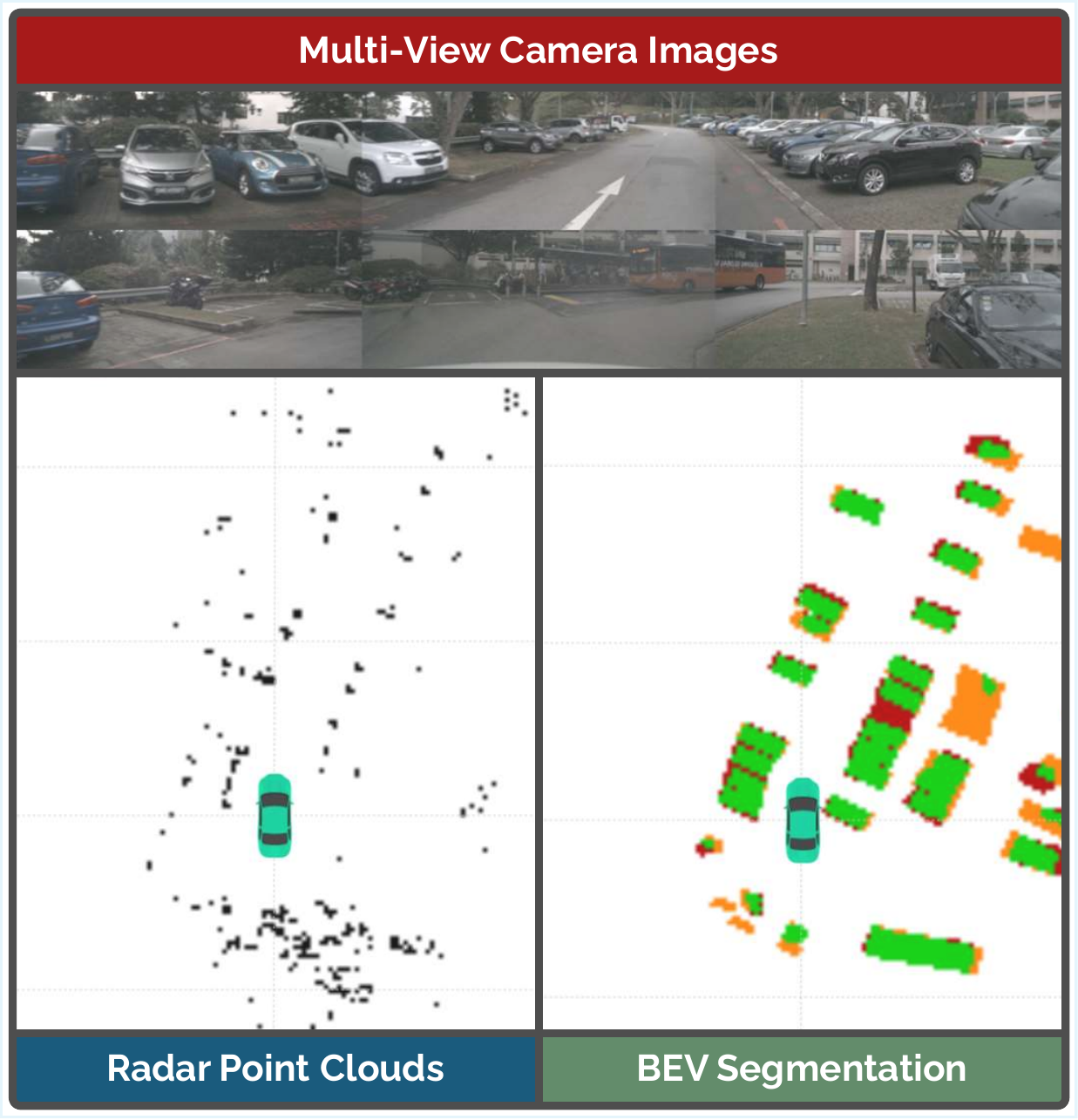}
  \caption{Our proposed method, \textbf{CaR1}, uses \textcolor[RGB]{158,0,0}{multi-view camera images} and \textcolor[RGB]{0,73,111}{radar point clouds} to predict a BEV vehicle segmentation map. Predictions are shown by colour: \textcolor{Green}{correct}, \textcolor{orange}{missing} and \textcolor{BrickRed}{wrong}.}
  \label{fig:motivation}
  \vspace{-0.5cm}
\end{figure}

In summary, we make three key claims: (i) our proposed \textbf{CaR1} achieves competitive results on BEV vehicle segmentation (57.6 IoU), on par with the state-of-the-art (SOTA) methods; (ii) our grid-wise feature encoding for radar point clouds integrates the strengths of a point-based network into a BEV framework; (iii) our adaptive fusion strategy effectively integrates the complementary advantages of camera and radar.

\begin{figure*}
    \centering
    \includegraphics[width=0.95\textwidth]{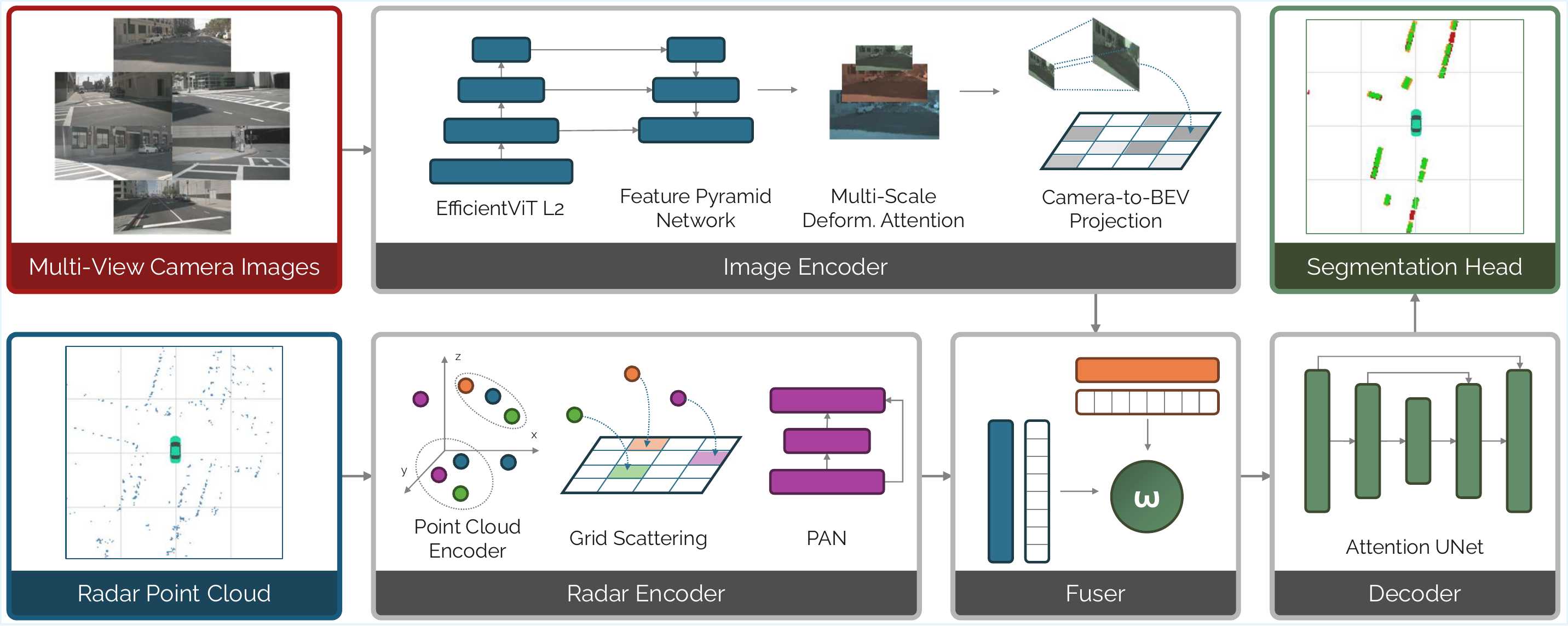}
    \caption{\textbf{Main diagram of our proposal, CaR1.} Using \textcolor[RGB]{158,0,0}{multi-view camera images} and \textcolor[RGB]{0,73,111}{radar point clouds}, we predict \textcolor[RGB]{82,128,91}{BEV segmentation maps}. We use two independent encoder branches, one for each modality, and we integrate features in an unified BEV space. We refine and decode the BEV feature using an Attention U-Net before a segmentation head.}
    \label{fig:architecture}
\end{figure*}

\pagebreak

\section{Related Work}

\noindent \textbf{Camera-based methods.} Estimating depth from a monocular camera is an ill-posed problem. Bridging the view disparity and lifting features to 3D space is a fundamental problem for robotics perception. Geometric or forward-based methods project learnt perspective features into BEV using a set of discrete pre-computed bins \cite{philion2020lift,liu2023bevfusion,li2023bevdepth}. Backward-based methods employ attention or interpolation to map correspondences from image space to feature space \cite{li2022bevformer,harley2023simple,man2023bev}. Recent frameworks combine both strategies for occupancy prediction and temporal detection \cite{li2023occ,li2024bevnext}.\\

\noindent \textbf{Radar-based methods.} Radar-only perception is less developed due to sparse measurements in conventional radars and limited datasets. nuScenes \cite{caesar2020nuscenes} provided the first benchmark with automotive radar, while newer datasets include next-gen sensors \cite{palffy2022multi,fent2024man}, though tasks such as BEV segmentation remain underexplored. Early solutions adapted LiDAR-based architectures to radar data \cite{lang2019pointpillars} Recently, radar-specific designs have appeared \cite{jia2025radarnext}.\\

\noindent \textbf{Fusion methods.} Radar provides spatial localization and motion cues, mitigating the depth ambiguity and temporal limitations of camera-only perception. SimpleBEV \cite{harley2023simple} fuses radar point clouds with lifted camera features via bilinear sampling. BEVCar \cite{schramm2024bevcar} integrates radar in the lifting process and applies attention-based fusion in BEV space. CRN \cite{kim2023crn} introduces dual-stage fusion, where radar occupancy guides camera-to-BEV lifting through depth-based cost volumes, followed by BEV fusion using cross-deformable attention. CRT-Fusion \cite{kim2024crt} applies dual-stage fusion by segmenting images into column tokens and by attending to azimuth-aligned radar tokens. In summary, fusion strategies combine radar with camera via three mechanisms: refining depth estimation, guiding the perspective-to-BEV transformation, and enabling spatially aware BEV-space fusion.
\section{Methodology} \label{sec:method}

Our goal is to perform accurate BEV vehicle segmentation using a set of images from a multi-camera system and radar point clouds, as depicted in Fig. \ref{fig:architecture}. Given the sensor data and their corresponding calibration matrices, \textbf{CaR1} predicts BEV segmentation maps, $S \in \mathbb{R}^{\left| \mathcal{C} \right| \times H_{BEV} \times W_{BEV}}$, where $\left| \mathcal{C} \right|$ is the number of classes, and $H_{BEV}$, $W_{BEV}$ are the dimensions of the BEV grid.

\subsection{Image Feature Extraction}

In this stage, multi-camera images $I$ are processed to generate a BEV feature map $F_{I,BEV}$. We adopt EfficientViT-L2 \cite{cai2023efficientvit}, a lightweight multi-scale Transformer, to extract ×8, ×16, and ×32 downsampled perspective feature maps. These are processed through an FPN \cite{lin2017feature} to yield $C=256$ channels and a Multi-Scale Deformable Attention (MSDA) module, following a Deformable-DETR \cite{zhu2020deformable} design, enabling self-attention across feature maps. Perspective features are then transformed into BEV space by projecting the centers of a predefined 3D voxel grid onto image planes using camera intrinsics and extrinsics. Valid projections falling within both the image plane and field of view are sampled via bilinear interpolation, while features from overlapping cameras are averaged. The resulting voxel-aligned features are reshaped into a BEV map, with the vertical dimension collapsed into channels.

\subsection{Radar Feature Extraction}

In this stage, the radar point cloud $R$ is processed to generate a BEV feature representation $F_{R,BEV}$ for fusion with camera features. We adopt a small Point Transformer V3 (PTv3) \cite{wu2024point} as a point-based encoder, applying serialized attention on the ordered point cloud. The PTv3 follows a U-Net design with a Transformer-based encoder and decoder. Point features are then projected to the BEV grid by computing each point’s $(x, y)$ coordinates and aggregating via differentiable scatter addition. To enhance the global context, a pyramid aggregation network is then applied, performing multi-scale fusion of the scattered radar features.

\subsection{Adaptive Feature Fusion}
We perform sensor fusion in the unified BEV space with an adaptive feature fusion mechanism that allows the combination of different sensor modalities with varying dimension. For each input modality, a double convolutional block maps the inputs to a common channel dimension. Then, an attention-based weighting is applied to each modality, dinamically adjusting their relative importance. Finally, a Squeeze-and-Excite (SE) operation performs a channel-wise recalibration on the weighted sum of features, according to Eq. \ref{eq:fuse}.
\begin{equation} \label{eq:fuse}
    F_{BEV} = SE\left( \sum_{i\:\in\:[I,R]} \text{ConvBlock}(F_{i,BEV}) \cdot \omega_i \right)
\end{equation}

\subsection{BEV Decoder}
We employ Attention U-Net \cite{oktay2018attention} as the BEV decoder to refine fused BEV features. It follows a standard U-Net architecture with attention blocks at the skip connections. At each decoder stage, attention is computed as
\begin{align*}
    x_d &= x_{d-1} \cdot \psi \\
    \psi &= \sigma(\text{Conv}(\text{ReLU}(\text{Conv}(x_{d-1})+\text{Conv}(x_e))))
\end{align*}
where $\psi$ denotes the attention weights, $\sigma$ is the sigmoid function, $x_d$ is the current decoder feature map, $x_{d-1}$ is the previous decoder map, and $x_e$ is the corresponding encoder feature map from the skip connection.

\subsection{Segmentation Head}
The BEV segmentation head first applies a differentiable bilinear grid sampling to transform features from pixel space to metric BEV coordinates. A double convolutional block then maps the features from $C=256$ to the number of output classes $M$; for vehicle segmentation, $M=1$.

\subsection{Loss Functions}
For BEV vehicle segmentation, ground-truth annotations are obtained by projecting all vehicle-like bounding boxes into a unified vehicle category. Supervision is provided via a combination of binary cross-entropy (BCE) loss for both the auxiliary and output heads, and a Dice coefficient loss applied to the output head. This combination ensures both per-pixel and region-level supervision. The total loss is defined as:
\begin{equation}
    \mathcal{L} = \alpha_1 \cdot \mathcal{L}_{BCE} + \alpha_2 \cdot \mathcal{L}_{BCE aux} + \alpha_3 \cdot \mathcal{L}_{Dice}
\end{equation}

with experimental weights set to $\alpha_1 = \alpha_3 = 1$ and $\alpha_2 = 0.4$.
\section{Experimental Evaluation}

\subsection{Dataset and Metrics}
We conduct experiments on the nuScenes dataset. The dataset contains 1,000 20-second driving scenes, split into 700 training, 150 validation, and 150 test sequences. The data is annotated with 3D bounding boxes to support object detection and tracking tasks. We quantify the performance of our proposal in the BEV vehicle segmentation task using Intersection over Union (IoU) or Jaccard index metric.

\subsection{Implementation Details}

\noindent \textbf{Optimization.} We train the model in an end-to-end manner for 40 epochs on a distributed setup of 2x Nvidia A100 80GB GPUs using DDP and gradient accumulation to achieve an effective batch size of 32. The AdamW optimizer is employed with a one-cycle cosine annealing learning rate scheduler (one-epoch warm-up, peak learning rate at $3\mathrm{e}{-4}$, decaying to zero) and weight decay of $0.01$. The architecture and codebase are implemented on PyTorch 2.4.1 and Lightning 2.4.\\

\noindent \textbf{Data Preprocessing and Augmentations}. 
Input images are resized to 320×800 and augmented via random horizontal flip, zoom, and rotation. Seven radar sweeps are accumulated and encoded using ordinal or one-hot representations. BEV-space augmentations, including rotation, scaling, and flipping, are applied to both inputs and ground-truth labels.\\

\noindent \textbf{BEV Grid Size}.
Experiments are performed in a range that spans 100 meters in both directions (from -50 to 50 meters), with the vehicle situated at the coordinate origin. The space is discretized with a grid resolution of 50 cm, resulting in a BEV grid size of 200 × 200 cells. Camera frustums are defined with depth bins from 1-60m at 0.5 m resolution. Internally, the network operates on a 256 × 256 BEV grid with 0.4 m resolution, which is resampled after the segmentation head.

\subsection{Results on BEV Vehicle Segmentation}

\begin{table}[]
\centering
\caption{BEV Vehicle Segmentation on the nuScenes Validation Set} \label{tab:sota}
\begin{tabular}{@{}lccccc}
\toprule
\multicolumn{1}{c}{\textbf{Method}} & \textbf{Modes} & \textbf{Cam Enc} & \textbf{Radar Enc} &\textbf{IoU} \\ \midrule
FIERY Static \cite{hu2021fiery}    & \multicolumn{1}{c}{C}   & \multicolumn{1}{c}{EN-b4}   & \multicolumn{1}{c}{-}   & \multicolumn{1}{c}{35.8} \\
BEVFormer \cite{li2022bevformer}       & \multicolumn{1}{c}{C}   & \multicolumn{1}{c}{RN-101}   & \multicolumn{1}{c}{-}   & \multicolumn{1}{c}{43.2}      \\
SimpleBEV \cite{harley2023simple}       & \multicolumn{1}{c}{C}   & \multicolumn{1}{c}{RN-101}   & \multicolumn{1}{c}{-}   & \multicolumn{1}{c}{47.4}      \\ \midrule
SimpleBEV++ \cite{harley2023simple}     & \multicolumn{1}{c}{C+R} & \multicolumn{1}{c}{RN-101}   & \multicolumn{1}{c}{PFE+Conv}   & \multicolumn{1}{c}{52.7}      \\
SimpleBEV \cite{harley2023simple}     & \multicolumn{1}{c}{C+R} & \multicolumn{1}{c}{RN-101}   & \multicolumn{1}{c}{Conv}   & \multicolumn{1}{c}{55.7}      \\
BEVCar \cite{schramm2024bevcar}    & \multicolumn{1}{c}{C+R} & \multicolumn{1}{c}{ViT-B}   & \multicolumn{1}{c}{PFE+Conv}   & \multicolumn{1}{c}{58.4}      \\
CRN \cite{kim2023crn}            & \multicolumn{1}{c}{C+R} & \multicolumn{1}{c}{RN-50}   & \multicolumn{1}{c}{SECOND}   & \multicolumn{1}{c}{58.8}      \\
BEVGuide \cite{man2023bev}              & \multicolumn{1}{c}{C+R} & \multicolumn{1}{c}{EN-b4}   & \multicolumn{1}{c}{SECOND}   & \multicolumn{1}{c}{\textbf{59.2}}      \\ \midrule \rowcolor{olive!12}
\textbf{CaR1 \textit{(ours)}}              & \multicolumn{1}{c}{C+R} & \multicolumn{1}{c}{EViT-L2}   & \multicolumn{1}{c}{PTv3+PAN}   & \multicolumn{1}{c}{57.6}      \\ \bottomrule
\end{tabular}
\end{table}

We evaluate our method on the nuScenes validation split and compare it with state-of-the-art approaches. Following \cite{harley2023simple,schramm2024bevcar}, vehicles with less than $40\%$ visibility are excluded, and predictions are thresholded at 0.5 with a batch size of 1. Results in Table \ref{tab:sota} show that our method outperforms all camera-only architectures by +10.2 IoU, demonstrating the benefit of incorporating radar measurements. Additionally, we achieve competitive performance relative to other SOTA methods, surpassing SimpleBEV by +1.9 IoU.

\subsection{Ablation Study}

\begin{table}[]
\caption{Ablation Study} \label{tab:ablation}
\centering
\begin{tabular}{@{}lcc@{}}
\toprule
\multicolumn{1}{c}{\textbf{Method}} & \hspace{0.2cm}\textbf{$\Delta$IoU}\hspace{0.2cm} & \hspace{0.3cm}\textbf{IoU}\hspace{0.3cm} \\ \midrule
Baseline & - & 35.0 \\
+ \textit{Higher Image Res.} & +0.6 & 35.6 \\
+ \textit{Swin-T} $\xrightarrow[]{}$ \textit{EViT L2} & +7.8 & 43.4 \\
+ \textit{Image Encoder Self Attn.} & +0.7 & 44.1 \\
+ \textit{PTv3 + Conv. Radar Encoder} & +8.0 & 52.1 \\
+ \textit{SECOND} $\xrightarrow[]{}$ \textit{Attn U-Net} & +4.0 & 56.1 \\
+ \textit{Data Aug. (Image + BEV)} & +1.5  & 57.6 \\
\bottomrule
\end{tabular}
\end{table}

We design our network by iteratively building on BEVFusion \cite{liu2023bevfusion}, originally developed for LiDAR–camera fusion. The baseline consists of (1) a Swin Transformer Tiny with FPN as the image encoder, (2) a SECOND-based BEV decoder, and (3) a grid sampling-based BEV segmentation head, with the PointPillars LiDAR encoder removed.

Our improvements include: (1) increasing image resolution from $(256, 704)$ to $(320, 800)$, (2) adopting EfficientViT L2 as a stronger image encoder, (3) adding self-attention in the image encoder, (4) incorporating a radar encoder with fusion modules, (5) replacing the BEV decoder with Attention U-Net, and (6) applying data augmentation in image and BEV spaces. Collectively, these enhancements improved BEV vehicle segmentation performance from $35.6$ to $57.6$ IoU, representing an increase of $+22.0$ (Table \ref{tab:ablation}).
\section{Conclusion}

In this report, we propose \textbf{CaR1}, a method for BEV vehicle segmentation in autonomous driving that fuses multi-camera images and sparse radar point clouds to robustly capture vehicle shapes in BEV perspective. By leveraging a grid-wise feature encoding of radar point clouds and an adaptive fusion strategy, we advance camera-radar fusion. Taking advantage of the complementary strengths of both sensors, our approach improves boundary delineation and segmentation accuracy. Evaluation on the nuScenes validation split demonstrates the effectiveness of our design, with the ablation study highlighting the contribution of each proposed improvement.


\footnotesize
\bibliographystyle{IEEEtran}
\bibliography{references.bib}

\begin{thebibliography}{10}
\providecommand{\url}[1]{#1}
\csname url@rmstyle\endcsname
\providecommand{\newblock}{\relax}
\providecommand{\bibinfo}[2]{#2}
\providecommand\BIBentrySTDinterwordspacing{\spaceskip=0pt\relax}
\providecommand\BIBentryALTinterwordstretchfactor{4}
\providecommand\BIBentryALTinterwordspacing{\spaceskip=\fontdimen2\font plus
\BIBentryALTinterwordstretchfactor\fontdimen3\font minus \fontdimen4\font\relax}
\providecommand\BIBforeignlanguage[2]{{%
\expandafter\ifx\csname l@#1\endcsname\relax
\typeout{** WARNING: IEEEtran.bst: No hyphenation pattern has been}%
\typeout{** loaded for the language `#1'. Using the pattern for}%
\typeout{** the default language instead.}%
\else
\language=\csname l@#1\endcsname
\fi
#2}}

\bibitem{caesar2020nuscenes}
H.~Caesar, V.~Bankiti, A.~H. Lang, S.~Vora, V.~E. Liong, Q.~Xu, A.~Krishnan, Y.~Pan, G.~Baldan, and O.~Beijbom, ``nuscenes: A multimodal dataset for autonomous driving,'' in \emph{Proceedings of the IEEE/CVF conference on computer vision and pattern recognition}, 2020, pp. 11\,621--11\,631.

\bibitem{philion2020lift}
J.~Philion and S.~Fidler, ``Lift, splat, shoot: Encoding images from arbitrary camera rigs by implicitly unprojecting to 3d,'' in \emph{Computer Vision--ECCV 2020: 16th European Conference, Glasgow, UK, August 23--28, 2020, Proceedings, Part XIV 16}.\hskip 1em plus 0.5em minus 0.4em\relax Springer, 2020, pp. 194--210.

\bibitem{liu2023bevfusion}
Z.~Liu, H.~Tang, A.~Amini, X.~Yang, H.~Mao, D.~L. Rus, and S.~Han, ``Bevfusion: Multi-task multi-sensor fusion with unified bird's-eye view representation,'' in \emph{2023 IEEE international conference on robotics and automation (ICRA)}.\hskip 1em plus 0.5em minus 0.4em\relax IEEE, 2023, pp. 2774--2781.

\bibitem{li2023bevdepth}
Y.~Li, Z.~Ge, G.~Yu, J.~Yang, Z.~Wang, Y.~Shi, J.~Sun, and Z.~Li, ``Bevdepth: Acquisition of reliable depth for multi-view 3d object detection,'' in \emph{Proceedings of the AAAI Conference on Artificial Intelligence}, vol.~37, no.~2, 2023, pp. 1477--1485.

\bibitem{li2022bevformer}
Z.~Li, W.~Wang, H.~Li, E.~Xie, C.~Sima, T.~Lu, Y.~Qiao, and J.~Dai, ``Bevformer: Learning bird’s-eye-view representation from multi-camera images via spatiotemporal transformers,'' in \emph{European conference on computer vision}.\hskip 1em plus 0.5em minus 0.4em\relax Springer, 2022, pp. 1--18.

\bibitem{harley2023simple}
A.~W. Harley, Z.~Fang, J.~Li, R.~Ambrus, and K.~Fragkiadaki, ``Simple-bev: What really matters for multi-sensor bev perception?'' in \emph{2023 IEEE International Conference on Robotics and Automation (ICRA)}.\hskip 1em plus 0.5em minus 0.4em\relax IEEE, 2023, pp. 2759--2765.

\bibitem{man2023bev}
Y.~Man, L.-Y. Gui, and Y.-X. Wang, ``Bev-guided multi-modality fusion for driving perception,'' in \emph{Proceedings of the IEEE/CVF Conference on Computer Vision and Pattern Recognition}, 2023, pp. 21\,960--21\,969.

\bibitem{li2023occ}
Z.~Li, Z.~Yu, D.~Austin, M.~Fang, S.~Lan, J.~Kautz, and J.~M. Alvarez, ``Fb-occ: 3d occupancy prediction based on forward-backward view transformation,'' \emph{arXiv preprint arXiv:2307.01492}, 2023.

\bibitem{li2024bevnext}
Z.~Li, S.~Lan, J.~M. Alvarez, and Z.~Wu, ``Bevnext: Reviving dense bev frameworks for 3d object detection,'' in \emph{Proceedings of the IEEE/CVF Conference on Computer Vision and Pattern Recognition}, 2024, pp. 20\,113--20\,123.

\bibitem{palffy2022multi}
A.~Palffy, E.~Pool, S.~Baratam, J.~F. Kooij, and D.~M. Gavrila, ``Multi-class road user detection with 3+ 1d radar in the view-of-delft dataset,'' \emph{IEEE Robotics and Automation Letters}, vol.~7, no.~2, pp. 4961--4968, 2022.

\bibitem{fent2024man}
F.~Fent, F.~Kuttenreich, F.~Ruch, F.~Rizwin, S.~Juergens, L.~Lechermann, C.~Nissler, A.~Perl, U.~Voll, M.~Yan, \emph{et~al.}, ``Man truckscenes: A multimodal dataset for autonomous trucking in diverse conditions,'' \emph{arXiv preprint arXiv:2407.07462}, 2024.

\bibitem{lang2019pointpillars}
A.~H. Lang, S.~Vora, H.~Caesar, L.~Zhou, J.~Yang, and O.~Beijbom, ``Pointpillars: Fast encoders for object detection from point clouds,'' in \emph{Proceedings of the IEEE/CVF conference on computer vision and pattern recognition}, 2019, pp. 12\,697--12\,705.

\bibitem{jia2025radarnext}
L.~Jia, R.~Guan, H.~Zhao, Q.~Zhao, K.~L. Man, J.~Smith, L.~Yu, and Y.~Yue, ``Radarnext: Real-time and reliable 3d object detector based on 4d mmwave imaging radar,'' \emph{arXiv preprint arXiv:2501.02314}, 2025.

\bibitem{schramm2024bevcar}
J.~Schramm, N.~V{\"o}disch, K.~Petek, B.~R. Kiran, S.~Yogamani, W.~Burgard, and A.~Valada, ``Bevcar: Camera-radar fusion for bev map and object segmentation,'' \emph{arXiv preprint arXiv:2403.11761}, 2024.

\bibitem{kim2023crn}
Y.~Kim, J.~Shin, S.~Kim, I.-J. Lee, J.~W. Choi, and D.~Kum, ``Crn: Camera radar net for accurate, robust, efficient 3d perception,'' in \emph{Proceedings of the IEEE/CVF International Conference on Computer Vision}, 2023, pp. 17\,615--17\,626.

\bibitem{kim2024crt}
J.~Kim, M.~Seong, and J.~W. Choi, ``Crt-fusion: Camera, radar, temporal fusion using motion information for 3d object detection,'' \emph{arXiv preprint arXiv:2411.03013}, 2024.

\bibitem{cai2023efficientvit}
H.~Cai, J.~Li, M.~Hu, C.~Gan, and S.~Han, ``Efficientvit: Lightweight multi-scale attention for high-resolution dense prediction,'' in \emph{Proceedings of the IEEE/CVF International Conference on Computer Vision}, 2023, pp. 17\,302--17\,313.

\bibitem{lin2017feature}
T.-Y. Lin, P.~Doll{\'a}r, R.~Girshick, K.~He, B.~Hariharan, and S.~Belongie, ``Feature pyramid networks for object detection,'' in \emph{Proceedings of the IEEE conference on computer vision and pattern recognition}, 2017, pp. 2117--2125.

\bibitem{zhu2020deformable}
X.~Zhu, W.~Su, L.~Lu, B.~Li, X.~Wang, and J.~Dai, ``Deformable detr: Deformable transformers for end-to-end object detection,'' \emph{arXiv preprint arXiv:2010.04159}, 2020.

\bibitem{wu2024point}
X.~Wu, L.~Jiang, P.-S. Wang, Z.~Liu, X.~Liu, Y.~Qiao, W.~Ouyang, T.~He, and H.~Zhao, ``Point transformer v3: Simpler faster stronger,'' in \emph{Proceedings of the IEEE/CVF Conference on Computer Vision and Pattern Recognition}, 2024, pp. 4840--4851.

\bibitem{oktay2018attention}
O.~Oktay, J.~Schlemper, L.~L. Folgoc, M.~Lee, M.~Heinrich, K.~Misawa, K.~Mori, S.~McDonagh, N.~Y. Hammerla, B.~Kainz, \emph{et~al.}, ``Attention u-net: Learning where to look for the pancreas,'' \emph{arXiv preprint arXiv:1804.03999}, 2018.

\bibitem{hu2021fiery}
A.~Hu, Z.~Murez, N.~Mohan, S.~Dudas, J.~Hawke, V.~Badrinarayanan, R.~Cipolla, and A.~Kendall, ``Fiery: Future instance prediction in bird's-eye view from surround monocular cameras,'' in \emph{Proceedings of the IEEE/CVF International Conference on Computer Vision}, 2021, pp. 15\,273--15\,282.

\end{thebibliography}


\end{document}